\documentclass[journal]{journal}
\ifCLASSINFOpdf
   \usepackage[pdftex]{graphicx}
   \graphicspath{{../pdf/}{../jpeg/}}
   \DeclareGraphicsExtensions{.pdf,.jpeg,.png}
\else
   \usepackage[dvips]{graphicx}
   \graphicspath{{../eps/}}
   \DeclareGraphicsExtensions{.eps}
\fi

\usepackage{mdwmath}
\usepackage{mdwtab}

\usepackage{caption}
\begin{document}
%
\title{Feature Extraction via Recurrent Random Deep Ensembles and its Application in Gruop-level Happiness Estimation}
%
%
%

\author{Shitao~TANG,
		Yichen~PAN
\thanks{Y. PAN and S. TANG are with the Computer Science Department, The University of Nottingham Ningbo China (corresponding authors, email: zy14897@nottingham.edu.cn, zy15779@nottingham.edu.cn). The order of author names is based on the alphabetical order of the first name. }
}

%
%

\markboth{Journal of \LaTeX\ Class Files,~Vol.~6, No.~1, January~2007}%
{Shell \MakeLowercase{\textit{et al.}}: Bare Demo of IEEEtran.cls for Journals}
%



\maketitle
\thispagestyle{empty}

\begin{abstract}
This paper presents a novel ensemble framework to extract highly discriminative feature representation of image and its application for group-level happpiness intensity prediction in wild. In order to generate enough diversity of decisions, n convolutional neural networks are trained by bootstrapping the training set and extract n features for each image from them. A recurrent neural network (RNN) is then used to remember which network extracts better feature and generate the final feature representation for one individual image. Several group emotion models (GEM) are used to aggregate face features in a group and use parameter-optimized support vector regressor (SVR) to get the final results. Through extensive experiments, the great effectiveness of the proposed recurrent random deep ensembles (RRDE) is demonstrated in both structural and decisional ways. The best result yields a 0.55 root-mean-square error (RMSE) on validation set of HAPPEI dataset, significantly better than the baseline of 0.78.
\end{abstract}


%
\IEEEpeerreviewmaketitle

\section{Introduction}
%
%
%
%

Groups are emotional entities are a rich source of varied manifestations of affect. Automatic Group-level Emotion Analysis (GEA) in images is an important problem as it has a wide variety of applications. In e-learning applications, group-level affective computing can be used to adjust the presentation style of a computerized tutor when a learner is bored, interested, frustrated, or pleased. In terms of social monitoring, for example, a car can monitor the emotion of all occupants and engage in additional safety measures, such as alerting other vehicles if it detects the driver to be angry. With regard to management of images, dividing images into albums based on facial emtion is a good solution to searching, browsing and managing images in multi-media systems \cite{dhall2010facial}. It also has pratical mearning in key-frame detection and event detection \cite{vandal2015event}.

Deep learning based approaches, particularly those using CNNs, have been very successful at image-related tasks in recent years, due to their ability to extract good representations from data. Judging a person's emotion can sometimes be difficult even for humans, due to subtle differences in expressions between the more nuanced emotions (such as sadness and fear). As a result, finely-tuned and optimized extracted features from images are of great importance in order for a classifier to make good predictions.

This paper demonstrates the efficacy of the proposed recurrent random deep ensembles on Group level emotion recognition sub-challenge based on HAPPEI database. The task is to infer the happiness intensity of the group as a whole on a scale from 0 to 5 from a bottom-up prespective. First CNN ensembles are introduced to extract several efficiently representative features of each face. Then feature aggrengation is conducted on each face using a Long Short Term Memory (LSTM). The proposed method will selectively memorize (or forget) the components of features which are important (or less important). Then face-level estimation is conducted using the trained SVR on the compact feature representation of each face. Various group emotion models are explored, including mean encoding and weighted fusion framework based on top-down features, such as sizes of faces and the distances between them. Note that the proposed method caters to aggregating information from multiple sources including different number of faces in one image. Our best result exhibits a RMSE of 0.55 on the validation set of HAPPEI dataset, which compares favorably to the RMSE of 0.78 of the baseline.

The rest of the paper is organized as follows. Section 2 describes related previous work. Section 3 describes in detail the proposed recurrent random deep ensembles for extracting facial features. Section 4 discusses the experiments and evaluates the results, and section 5 concludes the paper.

\section{Related Work}
The problem of emotion recognition in varied condition is multi-dimensional. There are two recent surveys which give detailed summary of related methodologies and the state-of-the-art in affect analysis \cite{corneanu2016survey}\cite{sariyanidi2015automatic}. In terms of the automatic affect recognition in diverse conditions, the EmotiW challenge series is an authoritative organization which aims at providing a platform for researchers to benchmark the performance of their methods on 'in the wild' data. Group-level Emotion Recognition (GReco) sub-challenge of EmotiW 2016 \cite{dhall2016emotiw} provides good benchmark for the research in GEA.

\subsection{Individual Facial Expression Recognition}	
Facial expression analysis usually employs a three-stage training consisting of feature learning, feature selection, and classifier construction. These features can be either hand-designed or learned from training images. Then, a subset of the extracted features, which is the most effective to distinguish one expression from the others, is selected to facilitate an efficient classification and enhance the generalization capability. Finally, a classifier is constructed given the extracted feature set for each target facial expression. Recently, it has been demonstrated that expression recognition can benefit from performing these three stages together with certain multi-layer (i.e., deep) neural network architectures. As a result, the majority of deep learning techniques applied for facial expression recognition ”in-the-wild” revolve around learning static discriminative templates via deep convolutional neural networks (DCNN) and using score aggregation for video classification to universal expressions \cite{kahou2013combining}.

Motivated by the success of the so-called multi-column DCNN (MCDNN) architecture \cite{ciregan2012multi} in various visual classification tasks, the MCDNN was applied for facial expression recognition ”in-the-wild” in \cite{kim2016hierarchical}. The standard MCDNN is a group of DCNNs with a simple averaging decision rule in a single structure level. Various network architectures, input normalization and random weight initialization were tested. Finally, in order to train more diverse decisions, an ensemble rule based on an exponentially-weighted decision fusion was applied. The best architecture achieved a recognition rate of around 57\% which was the highest reported in EmotiW 2015.

\cite{liu2014facial} proposes a novel Boosted Deep Belief Network (BDBN) for performing the three training stages iteratively in a unified loopy framework. More recent studies adopt CNN architectures that permit feature extraction and recognition in an end-to-end framework. For instance, \cite{yu2015image} employed an ensemble of multiple deep CNNs. \cite{mollahosseini2016going} used three inception structures proposed by \cite{szegedy2015going} in convolution for facial expression recognition. The Peak- Piloted Deep Network (PPDN) \cite{zhao2016peak} is introduced to implicitly learn the evolution from non-peak to peak expressions.

\subsection{Group-level Emotion Analysis}
Groups are emotional entities and a rich source of varied manifestations of affect. Social psychology studies suggest that group emotion can be conceptualized in different ways. Generally, group emotion analysis problems can be solved using two approaches. The firs is top-down approach, where emotion emerges at the group level and follows by individual participants of the group \cite{barsade1998group}. The second is bottom-up approach, where overall emotion of group is assumed to be constructed by uniqueness of individual members’ emotion expression \cite{kelly2001mood}.

\cite{dhall2015more} proposed a hybrid approach, which combines top-down and bottom-up components, where top-down information is extracted using scene descriptors and bottom-up information is extracted by analyzing the face of the members of a group. Later, \cite{huang2016analyzing} came up with a multi-modal method combining face, upperbody and context information, where face/upperbody is viewed as the bottom-up component and scene as top-down component. For representing an image, information aggregation was proposed to encode multiple people’s information for group-level image.

In the recent GReco Sub-challenge of EmotiW 2016, the baseline feature used is the CENsus TRansform hISTogram (CENTRIST) descriptor \cite{wu2011centrist}. \cite{li2016happiness} proposed a framework based on ensemble of features in LSTM and ordinal regression, which won the sub-challenge. The second place is the method from \cite{vonikakis2016group}, which is based on geometric features extracted from faces in an image. Partial least square regression is used to infer the group-level happiness intensity. \cite{sun2016lstm} proposed a LSTM based approach and fined tuned the AlexNet model \cite{krizhevsky2012imagenet}  by training on the Static Facial Expressions in the Wild (SFEW) dataset \cite{dhall2011static} and the HAPPEI dataset \cite{dhall2015automatic}.

\subsection{Dataset}
The dataset used in the experiment is the Happy People Images (HAPPEI) dataset \cite{dhall2015automatic}, which is a fully annotated dataset (by human labellers) for the task of group happiness level prediction. Every image in the data set is given a label between [0, 5]. The training and validation sets contain 2,836 images and 10,400 faces in total as shown in Table 1. However, the HAPPEI dataset is a highly unbalanced dataset, with most of the faces/groups labelled as 3, but very few of them labelled as 0 or 5. Training directly with such skewed data may compromise the generalization ability of the whole system.
\begin{table}
	\centering
	\begin{tabular}[t]{|l|l|l|}
		\hline Subset & Images & Faces \\
		\hline Traing set & 1500 &5549 \\
		\hline Validation set & 1138 & 4851 \\
		\hline
	\end{tabular}
	\caption*{Table 1: {\bf HAPPEI dataset}}
\end{table}

In this case, the existing dataset is resampled in order to create a more balanced subset, which is subsequently used as training set for face-level happiness estimation. More specifically, the aim is at creating a uniform training distribution, which includes all the training examples of the most under-represented intensity quantization bin: those annotated as 0 (neutral) and those annotated as 5 (thrilled). Finally, 380 training examples per intensity bin are selected, which is the number of faces annotated as 5. This results in a training subset with a total of 380 * 6 = 2280 training faces.

\section{Proposed Approach}
\subsection{Recurrent Random Deep Ensembles (RRDE)}
\cite{lee2016stochastic} uses stochastic multiple choice learning to train n networks and improve the accuracy of cifar-10 by 2-3\%. This method proves that CNN can be used as base classifier in ensemble learning. Inspired by this framework, an original feature extraction framework RRDE is proposed. It first trains n CNNs for each image, then extracts one vector representation from each network, and finally aggregates n feature vectors for each image into one compact feature vector using LSTM.

\subsection{Bootstrapping}
Although random weight initialization and Stochastic Gradient Descent(SGD) can also make diverse CNNs, these CNNs turn out to be very similar if they are trained with the same dataset. In this case, the performance is not gained very much when aggregating them together. This is one main cause of the relatively worse performance of CNN in ensemble learning. \cite{lee2016stochastic} uses stochastic multiple choice learning to address the problem. They select the network with the minimum loss and backward it, which makes the difference between CNNs more distinct. However, it is not effective when used in a relativley small dataset like HAPPEI dataset.

In order to make each CNN diverse, we adopt the first step of random forest, bootstrapping \cite{breiman2001random}. For every class, a random sample is selected repeatedly B times with replacement from the training set. After balancing the dataset, 380 images (the minimum number of images in a class) for each class are selected and for each network, bootstrapping is performed once so that each network is trained with different set of data (6*380=2280 images).

\subsection{LSTM based Feature Aggregation}
One straightforward way to aggregate the features is mean encoding. However, due to the diversity of these CNNs, for the same image, the difference between features extracted by different network can be huge. Extreme values have large impact on the final result. Therefore, mean encoding is not an ideal method for aggregation in this case. Inspired by the success of LSTM in sequencing learning \cite{sak2014long}, it is believed that LSTM can be used for aggregating these features. Illustration of one LSTM cell is in Figure 1. A cell memory C is used to memorize good feature representation of face feature vectors when scanning the sequence of feature vectors. The forget gate will decide how much of the content in the cell memory should be forgotten and the input gate in LSTM is employed to determine how much the new feature should be fed into the cell memory. The output gate decides how the features in the cell memory are converted to output features. The pipeline for LSTM based feature aggregation is shown in Figure 1.
\begin{figure}[h!]
	\centering
	\includegraphics[width=1.0\linewidth]{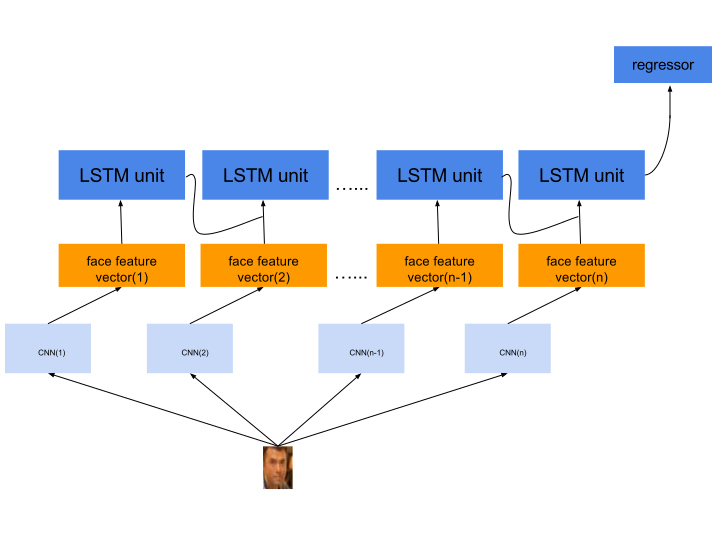}
	\caption{{\bf LSTM based feature aggregation pipeline}}
\end{figure}

\subsection{Group Emotion Model (GEM)}
After obtaining face-level happiness estimation of each individual face in an image, they need to be combined in order to get an estimation of the overall happiness of the group in image. In this subsection, four different Group Emotion Models for group-level estimation before feeding into the final regressor are discussed.

\subsubsection{Mean of Face-level Estimations}
One naive way for overall group-level happiness estimation is to use the mean of all face-level estimations in the group as introduced by \cite{dhall2015automatic}. 
\begin{equation}
g=\frac{\sum_{i=1}^{s}f_i}{s}
\end{equation}
where $g$ is the happiness intensity estimation of the group, $s$ is the total number of detected faces in this group and $f_i$ is the happiness intensity estimation for face $i$.

\subsubsection{Mean Encoding}
Another straightforward feature aggregation is mean encoding. The average of the face features is used as the feature representation of the whole image.

\subsubsection{Context-based weighted GEM} In above two GEMs, both global information, e.g. the relative position of people in the image, and local information, e.g. the level of occlusion of a face, are ignored. In this case, the effectiveness is compromised because all detected faces are assumed to contribute equally to the group-level estimation, which has been shown not generally the case \cite{dhall2015automatic}. In order to add the bottom-up and top-down components, the significance $s_i$ of a face $i$ is estimated using the following equation:
\begin{equation}
s_i=\frac{\Theta_i}{\delta_i}
\end{equation}
where $\Theta_i$ represents the level of occlusion of face $i$, and $\delta_i$ represents the relative position of the face $i$ among the group.

Based on the size of the face, $\Theta_i$ is further equal to the size of the bounding box of face $i$ (in pixels). Based on the positions of all faces in the group, $\delta_i$ is further equal to:
\begin{equation}
\delta_i=\sum_{j=1}^{n}||c_i - c_j||
\end{equation}
where $c_j$ and $c_i$ represent the coordinate of the centroid of corresponding face.

Equation (2) essentially adds social context features as weights to the process of determining the happiness intensity of a group image by normalizing the occlusion level of a face by the sum of its Euclidean distance from all other faces. In this case, small faces which are located away from the group are penalized, while larger faces which are closer to all others are assigned a higher significance. If n = 1, then significance is set to 1. In the experiment, this context-based weight scheme is further cooperated together with mean of face-level estimation and mean encoding GEM.

\section{Experiments}
In this section, results of the proposed RRDE based group-level happiness estimation are presented. For all the experiments in this section, the models are trained on the balanced training set of HAPPEI and tested on the validation set of HAPPEI.

\subsection{Facial Feature Extraction}
To estimate group happiness intensity, the most important step is to find good face feature representation. \cite{vonikakis2016group} uses the geometric feature of one face, which proves such feature is useful to measure the happiness intensity. However, this kind of feature is too hand-crafted to generalize on various qualities of face images. Convolution neural network (CNN) has showed its powerful capability to extract features \cite{krizhevsky2012imagenet}, and deep residual network is currently one of the state-of-art CNN structures for object recognition \cite{he2016deep}. Therefore, ResNet is adopted as the base feature extraction method. Since HAPPEI dataset is of relative small size, using deeper network like ResNet-50 or ResNet-101 will result in overfitting easily. After experiments, a 20-layer ResNet is finally trained, which has the same structure as the one trained by \cite{he2016deep} on Cifar-10 dataset. Cifar-10 dataset consists of 60000 images and 10 classes \cite{krizhevsky2009learning}. For each face image, it is forwarded through the whole network. The activations from the penultimate layer is extracted as face image representation. The dimension of the extracted face feature is 64.

A 20-layer ResNet is trained using the open source framework Tensorflow. The network inputs are 32*32*3 images, with every pixel value scaled to [-1,1]. A weight decay of 0.00001 is used. The initial learning rate is 0.01 and the batch size is 32. A CNN for 20000 iterations is trained and every 5000 iterations, learning rate is multiplied by 0.1. Measuring happiness is a regression problem, so it is supposed to use Squared Hinge Loss (L2-loss) in this problem, but since the interests are only in the feature extracted by CNN, cross entropy loss can also be used.  In the experiment, the feature extracted by the network with cross entropy loss are found to be better than L2-loss. Therefore cross entropy loss is used with 6 classes as the final loss layer. Bilinear interpolation is used to resize these images and randomly adjust the brightness, contrast, saturation when training so the data set is augmented to train a more robust network. In total, there are n networks and n feature representations for each image.

When evaluating the effectiveness of network ensembles, the majority voting strategy is used so that each image is predicted as the maximum number of labels it gets. The model under different number of networks is evaluated and it is found that there is no obvious improvement when using 6 or more networks. This may result from the fact that the ensembles tend to saturate quickly as the ensemble size grows. The result is shown in in Figure 2.
\begin{figure}[h!]
	\centering
	\includegraphics[width=1.0\linewidth]{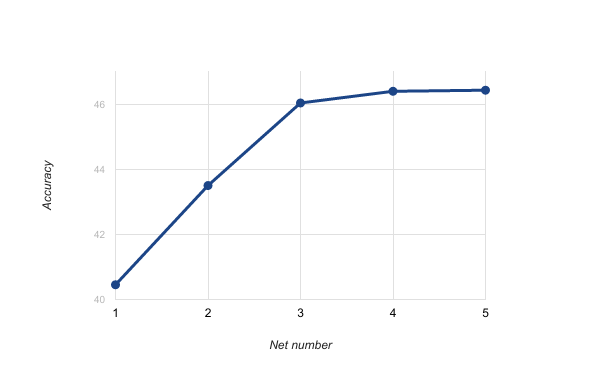}
	\caption{{\bf Evaluation of performance when using majority voting strategy}}
\end{figure}

\subsection{Facial Feature Aggregation}
LSTM is used to scan features extracted by CNN. 2 cells are stacked into the LSTM unit, and the memory size for each cell is 128. The order of scanning these features is the same so that LSTM can remember which network extracts better feature for a specific image. For every face image in the whole dataset, it is fed to n networks and get n 64-dimensional feature vectors. Then LSTM scans these vectors and L2-loss is used as loss function when training. Finally, a 128-dimensional vector is extracted from the final LSTM output for the final group happiness intensity analysis. RMSE is used to measure the performance of individual-level happiness intensity. 

In order to prove the effectiveness of the proposed RRDE method, a single ResNet is trained using all training data of HAPPEI with the L2-loss as loss function. All the parameters are the same as those CNNs in the ensembles except for the iteration changed from 20000 to 30000. The result is shown in Figure 3. This single ResNet is named as ResNet-All because it will be used again in the group-level happiness estimation comparison.
\begin{figure}[h!]
	\centering
	\includegraphics[width=1.0\linewidth]{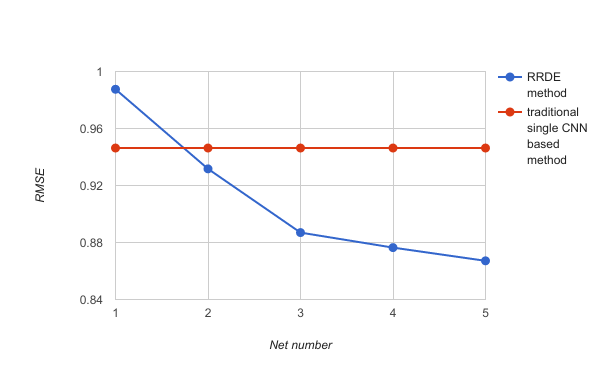}
	\caption{{\bf Comparison between the proposed RRDE method with the traditional single CNN based method on individual happiness estimation}: when the number of networks in the ensembles is 1, the performance is relatively worse than the single ResNet trained with all data. This may result from the fact that all data provides more information for the network to generalize. However, with the number of networks in the ensembles increases, the performance improves significantly.}
\end{figure}

\subsection{Group Emotion Modeling}
After getting one efficient feature representation for each face image, the effectiveness of four GEMs mentioned in Section 3 is tested. As shown in Table 2, for each of four GEMs, including normal mean-encoding, normal mean of face-level estimations, context-based weighted mean-encoding and mean of face-level estimations, the model with different features extracted from ensembles of different size is tested. 
\begin{figure}[h!]
	\centering
	\includegraphics[width=1.0\linewidth]{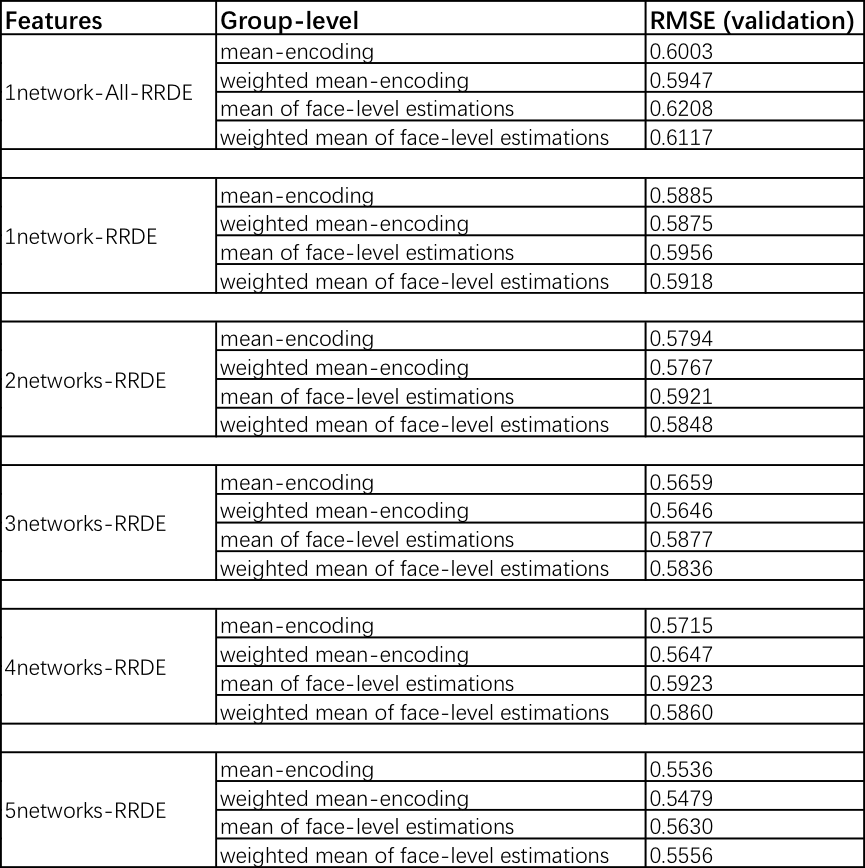}
	\caption*{Table 2: {\bf Results on the validation set}: 1network-All-RRDE refers to the features extracted from the RRDE with only one CNN trained on whole training set. The final regression is done by a SVR with penalty parameter C equal to 0.5, epsilon E equal to 0.13 and linear kernel type. The best result is {\bf0.5479} achieved by a 5networks-RRDE with weighted mean-encoding.}
\end{figure}

One immediate observation from Table 2 is that the performance of GEM improves consistently as the ensemble size increases. With the increase in the size of ensembles, more reliable and effective feature representation could come from the strong consensus of multiple sub-networks in the ensemble. Another insight from Table 2 is that the feature extracted from the ensembles trained after bootstrapping is much more effective than that trained by all training data.

Mean-encoding model consistently exhibits more competitive results than the simple averaging of face-level estimations, which indicates that the extracted feature contains very useful information, that may otherwise be ignored by simple averaging face-level estimations.

The significance factor of each face, which takes into consideration the relative position of people in the image, and the level of occlusion of a face, exhibits a modest performance improvement. When compared to the simple GEM without weighting scheme, the addition of the face significance factor improves the RMSE. Since no training method is involved in the use of a simple averaging, this improvement may not be the result of overfitting, but may have to do with the efficiency of the feature.

Moreover, the performance of the proposed RRDE method on group-level happiness estimation with the traditional single CNN based method is also compared. The same CNN called ResNet-All mentioned in Figure 3 is used here. Since the output vector from ResNet-All is of 64 dimension, the LSTM hidden units are resized from 128 to 64 so that the aggregated feature output from LSTM is of 64 dimension. The group-level estimation is based on mean-encoding GEM.
\begin{figure}[h!]
	\centering
	\includegraphics[width=1.0\linewidth]{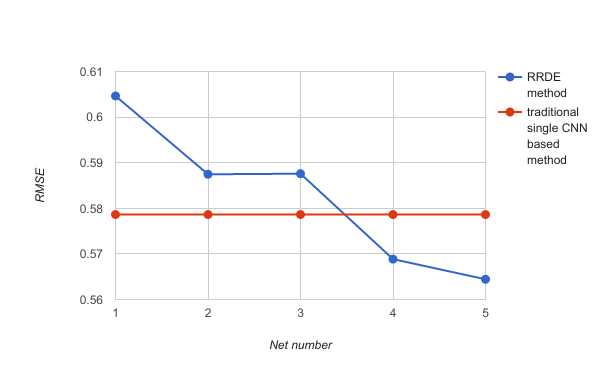}
	\caption{{\bf Comparison between the proposed RRDE method with the traditional single CNN based method on group-level happiness estimation}: The RMSE Similar to the case of individual-level happiness estimation, the proposed RRDE method beats the traditional method quickly with the number of networks in the ensemble increases.}
\end{figure}

\subsection{Regression}
The grid search combined with cross-validation is used to find the optimal parameters of SVR with penalty parameter C equal to 0.5, epsilon E equal to 0.13 and linear kernel type. The final SVRs for group-level estimation experiments mentioned above have the same hyper-parameters, but are trained on corresponding training sets. The details of the best results, together with the baseline provided by the challenge \cite{dhall2016emotiw}, are depicted in Table 3.
\begin{table}
	\centering
	\begin{tabular}[t]{|l|l|}
		\hline Model & RMSE \\
		\hline Baseline \cite{dhall2016emotiw} & 0.78 \\
		\hline 5-networks RRDE & 0.55 \\
		\hline
	\end{tabular}
	\caption*{Table 3: {\bf Final result}}
\end{table}

\section{Conclusion}
This paper proposes a new ensemble learning based method to estimate group-level happiness intensity. In total there are n networks with different data selected using bootstrapping and these networks are used to extract n 64-dimensional vectors for each image. LSTM is used to scan these vectors to aggregate them. The final output from LSTM is used as the individual face feature (128-dimensional vector). For both face-level and group-level estimation, a parameter-optimized SVR is used for regression. To aggregate group-level feature, four different GEMs including mean-encoding, weighted mean-encoding, mean of face-level estimations and weighted mean of face-level estimations are tried and get the best result is achieved when using weighted mean-encoding model with 5-network RRDE. This method can also be used for classification.


%

%

%
%

\ifCLASSOPTIONcaptionsoff
  \newpage
\fi



\bibliographystyle{IEEEtran}
\bibliography{main}
\end{document}